\documentclass[letterpaper, 10 pt, conference]{ieeeconf}  

\IEEEoverridecommandlockouts                              

\usepackage{amsmath,amssymb,amsfonts}
\usepackage{graphicx}
\usepackage{float}
\usepackage{textcomp}
\usepackage{xcolor}
\usepackage{gensymb}
\usepackage[english]{babel}
\usepackage[utf8]{inputenc}
\usepackage{array}
\usepackage{hyperref}
\usepackage{algorithm}
\usepackage{xcolor}
\usepackage{soul}
\usepackage[noend]{algpseudocode}
\usepackage[font=small,labelfont=bf]{caption}
\usepackage{subcaption}
\def\BibTeX{{\rm B\kern-.05em{\sc i\kern-.025em b}\kern-.08em
    T\kern-.1667em\lower.7ex\hbox{E}\kern-.125emX}}

\newcolumntype{M}[1]{>{\centering\arraybackslash}m{#1}}
\usepackage{etoolbox}
\makeatletter
\patchcmd{\@makecaption}
  {\scshape}
  {}
  {}
  {}
\makeatother
\begin{document}

\title{ An Efficient Method for Accurate Pose Estimation and Error Correction of Cuboidal Objects
\thanks{Utsav Rai, Hardik Mehta, Vismay Vakharia, Aditya Choudhary, Amit Parmar, Rolif Lima and Kaushik Das are with Embedded Systems and robotics, Tata Consultancy Services, Research and Innovation, Bengaluru, India.
}
}

\author{Utsav Rai, Hardik Mehta, Vismay Vakharia, Aditya Choudhary, Amit Parmar, Rolif Lima and  Kaushik Das}

\maketitle
\begin{abstract}

The proposed system outlined in this paper is a solution to a use case that requires autonomous picking of cuboidal objects from an organized or unorganized pile with high precision. This paper presents an efficient method for precise pose estimation of cuboid-shaped objects, which aims to reduce error in target pose in a time-efficient manner. Typical pose estimation methods like global point cloud registrations are prone to minor pose errors for which local registration algorithms are generally used to improve pose accuracy. However, due to the execution time overhead and uncertainty in the error of the final achieved pose, an alternate, linear time approach is proposed for pose error estimation and correction. This paper presents an overview of the solution followed by a detailed description of individual modules of the proposed algorithm.
\end{abstract}

Pose Estimation, Object Detection, 3D Matching, Pose Correction, Mobile manipulator, Palletization

\section{Introduction} \label{intro}
With the advance in industrial robots and sensors, the task of autonomous robotic picking and place is now more feasible than before. This improvement opens new opportunities in various use cases, including warehouse automation and bin-packing, autonomous-cleaning robots, rovers for planetary surface explorations and autonomous construction. One such application of mobile manipulation was presented in a use case comprised of a task involving picking different types of cuboidal objects with high precision in a time-constrained scenario. 

Usually, to determine a precise pose of an object, global point cloud registration methods are a good starting point as it estimates an approximate pose in less time, provided the object classification and segmentation are already done. But the resulting pose is often erroneous to a certain degree that must be addressed with correction method. Compensating for this error, the second stage of processing is usually done by local registration methods to achieve the desired accuracy. Methods such as Iterative closest point (ICP) \cite{zhang2021iterative} reduces the pose error by local registration method but are time consuming and still might be erroneous depending on the dataset. However, an outdoor environment involves various factors like acquiring data in varying lighting conditions, such as under direct sunlight or a poorly lit environment. In such scenarios majority of depth cameras capture erroneous point clouds primarily due to reflections or the inability to utilize colour properties in darker environments which needs to be compensated with data processing techniques and the use of depth sensors that can be tuned to work in flexible conditions.

Picking the correct object from the scene is a crucial step for its pose estimation that requires object identification and segmentation. Since the accuracy of the pose is essential for bin packing of the cuboidal objects, the solution must prevent these marginal pose errors in the minimum possible time. Hence we propose a method that uses the RGB data to segment the pointcloud on which global point cloud registration(Super4PCS \cite{mellado2014super}) is done. And to improve pose accuracy resulting from that are corrected by estimating the error in pose accurately using a point selection process and correcting the pose in linear time.



\section{Related Work}

\textit{Pose estimation using local registration:}
Methods such as Iterative closest point (ICP)\cite{zhang2021iterative} target scan pairs with small motion between them and are static. The algorithm with its worst-case running time between  $\Omega(n\log n)$ and $\mathcal{O}(n^{2}d)^{d}$ \cite{arthur2006worst} (where \textit{d} denotes dimension) is not an ideal time complexity for real-time pose estimation applications. Also, its convergence is guaranteed only when the scanned pair is roughly aligned. 

\textit{Pose estimation using global registration:}
The approach followed by \cite{mitash2018robust} merges the approach of the sampling-based registration technique with the global geometric modelling of objects to deliver results as good as the state-of-the-art Super4PCS \cite{mellado2014super} method. The algorithm proposes a slight improvement in accuracy when compared to Super4PCS but does not claim to be error-free. The average rotational and translation error by this approach was 6.29\degree and 1.11 cm, respectively, when applied over the Amazon Picking Challenge (APC) dataset .\par  
\textit{Pose estimation using both global and local registration:}
Another approach by \cite{lu2019point}, first aligns the point cloud data with global registration methods such as Super4PCS, but due to uncertainty in the accuracy of the resultant pose, ICP is used to align the data further to improve the pose accuracy. This approach trades off the execution time of ICP for reducing pose error, but in applications where both time and precision are of great importance, an alternate approach is required to deliver better execution time and accuracy in the pose.

\section{Proposed Approach} \label{section:soln}
The challenge involved in achieving efficient pose estimation and solving the problem statement of autonomous picking of cuboidal objects in an outdoor environment is given by two interchangeable approaches. The proposed approach uses RGB image to segment the cuboids' magnetic plate from the scene point cloud using aligned RGB and aligned depth image thus eliminating the need for any post-processing of raw point cloud. The pose estimation using Super4PCS, a global point cloud registration technique follows the segmentation process, and the resultant pose is then checked and corrected for any error in its rotation and translation in linear time. The overall solution presented remains faster and more accurate than the conventional approach involving local registration methods like ICP to correct pose deviations.

\begin{figure*}
   \centering
    \includegraphics [width=\textwidth]{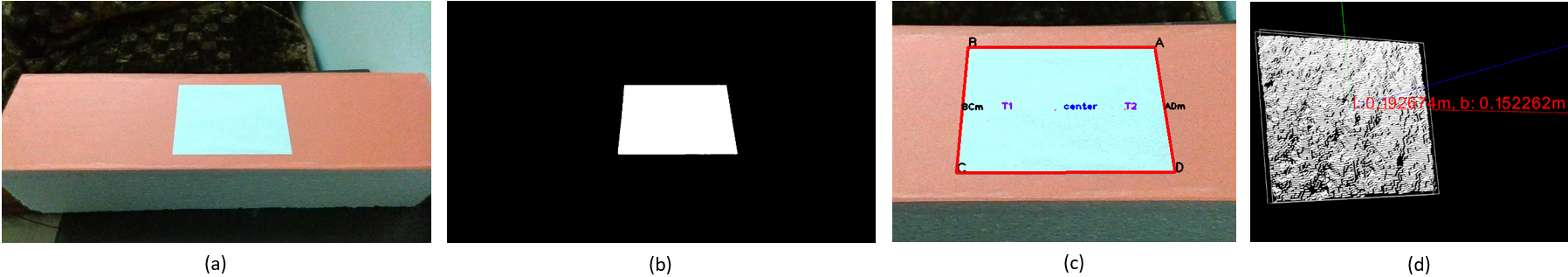}
    \caption{(a) RGB image of the scene aligned to the point cloud, (b) ROI thresholding to reduce search space, (c) T1, T2 and corner point selection and extraction of corresponding 3D point from point cloud using convexity defect and inverse projection, (d) Point cloud segmentation using non zero pixels from \textit{(b)} and fitting oriented bounding box to cross verify dimensions of ROI}
    \label{fig:rgb_process}
\end{figure*}

\subsection{Sensor Data Acquisition} \label{section:data_acq}
Only Intel Realsense D415 performed optimally in direct sunlight among the various depth cameras. With its field of view (FOV) of \((65^{\circ}\pm2^{\circ} \times 40^{\circ}\pm1^{\circ} \times 72^{\circ}\pm2^{\circ})\) and depth resolution up to \(1280 \times 720\) in both indoor and outdoor environment allowed it to be the ideal choice. The ability of D415 to deliver both RGB and depth images aligned to the point cloud (Fig \ref{fig:rgb_process}(a)), facilitates the proposed algorithm in reducing the execution time by decreasing the dataset significantly as only the points corresponding to the region of interest is processed instead of processing the complete scene point cloud.

\subsection{Sensor data processing} \label{section:data_process}
We have used the Point Cloud Library (PCL)\cite{rusu20113d} to process the point cloud with the following methods to remove any inconsistencies from the point cloud data and remove unnecessary information which is insignificant for the algorithm to deliver any meaningful result.\par 

\textit{Point cloud cropping:}
Since we are aware that any object in the scene is at a certain height from the ground, it is better to remove ground points as it does not contribute to pose estimation but adds to the time taken to process the point cloud. Point cloud cropping techniques such as \textit{Passthrough Filter} allow filtering out all the points along a specific axis after a specified distance. Other methods, such as convex and concave hull, allow irregular crop shapes from the point cloud.\par

\textit{Outlier Removal:} Varying lighting condition during data acquisition complicates the estimation of characteristics of local point clouds, such as curvature changes and surface normals, which leads to erroneous data causing registration failures. Hence, after the pass-through filter is applied, Statistical Outlier Removal \cite{balta2018fast} (SOR) method is applied to remove significantly smaller point clusters or individual points.

\textit{Downsampling:}
Due to higher resolution, it becomes necessary to downsample the point cloud. \textit{VoxelGrid} is a method that creates voxels (3D boxes of specified dimensions) across the dataset, and for each voxel, it approximates all the points to the centroid of the voxel. This method reduces the number of data points from the point cloud but preserves the structural information.\par

\textit{Smoothing and normal estimation:}\label{itm:mls}
SOR fails to remove data irregularities that are caused by small distance measurement errors and are difficult to remove using statistical analysis. To reduce any possibility of erroneous registration, \textit{re-sampling} algorithm such as \textit{Moving Least Square}\cite{kang2020point}(MLS) is applied, which uses higher-order polynomial interpolation between neighbouring points to recreate the missing parts of the surface, resulting in smoother planes.\par
    
\subsection{Segmentation} \label{section:segmentation}
\textit{Region Growing Plane Segmentation:}
Region Growing \cite{vo2015octree} is a segmentation algorithm that uses smoothness constraint and the angle between points normals to merge into a cluster of points that belong to the same smooth surface. The ROI segmentation using this approach solely depends on the surface normals; thus, no colour information is taken into account. Thus ROI filtration is done by using the dimensional characteristics of the target object.\par  


\textit{RGB based segmentation:}ROI segmentation using RGB image is done using HSV thresholding of the identified ROI in the source image (Fig. \ref{fig:rgb_process}(a)) and extracting the pixel locations of the ROI represented by non-zero pixel values as shown in Fig. \ref{fig:rgb_process} (b). The corresponding point cloud segment is then extracted using the non-zero pixel coordinates as indices for the input point cloud. This method allows the proposed algorithm to save time and computational overhead.

\subsection{ROI Filtration} \label{section:filter}
The resulting segments are checked for ROI using the target object's dimensional and colour properties. The measurement of segment's dimensions is done by fitting a \textit{Oriented Bounding Box (OBB)} over the segments which provides an approximate dimension of the rotated bounding box enclosing target object, as shown in Fig \ref{fig:rgb_process}(d). The OBB method computes eigenvectors from the covariance matrix of the segment point cloud, which is used to transform the cloud to the origin such that the principal component corresponds to the same axes and the maximum and minimum point of the transformed cloud is calculated. Finally, the rotation and translation of the bounding box are calculated using the eigenvectors to get an approximate bounding box for the segment, and the maximum and minimum points calculated are used to get box height, width and depth, as shown in Fig \ref{fig:rgb_process}(d).\par

Whereas the proposed algorithm calculates the dimension more accurately by fitting a quadrilateral over the segmented contour instead of using a minimum area rotated rectangle. Convexity defect allows fitting a polygon and helps calculate the dimension of ROI by using defect points as vertices, which are used as indices to retrieve corresponding 3D coordinates from the scene point cloud. Fig \ref{fig:rgb_process}(c), shows the least area quadrilateral fitted over the ROI.\par

The computed corners of the ROI are used for calculating the dimension of the segment by using the standart pinhole camera projection model. Using the camera model, inverse projection of pixel coordinate \(x,y\) is done to calculate \(X,Y\) world coordinate in camera frame by using the depth value at pixel coordinate \(x,y\) in aligned depth image denoted by \(Z\). The final equations giving the 3D coordinate in camera frame is stated as equation \ref{eq:inverse_proj_x}, \ref{eq:inverse_proj_y} and \ref{eq:inverse_proj_z}, in which camera's focal length \((f_x, f_y)\) and its principal point \((c_x, c_y)\) is derived from its intrinsic parameters as stated in equation \ref{eq:intrinsic_matrix}
\begin{equation} \label{eq:intrinsic_matrix}
K = \begin{bmatrix}
f_x & 0 & c_x \\
0  & f_y & c_y\\
0 & 0 & 1
\end{bmatrix}
\end{equation}

\begin{equation} \label{eq:inverse_proj_x}
X = (x - c_x) \times depth[y, x] / f_x 
\end{equation}
\vspace{-10pt}
\begin{equation} \label{eq:inverse_proj_y}
Y = (y - c_y) \times depth[y, x] / f_y 
\end{equation}
\begin{equation} \label{eq:inverse_proj_z}
Z = depth[y,x]
\end{equation}
\begin{figure*}[h!]
\centering
\includegraphics[width=0.85\textwidth, keepaspectratio]{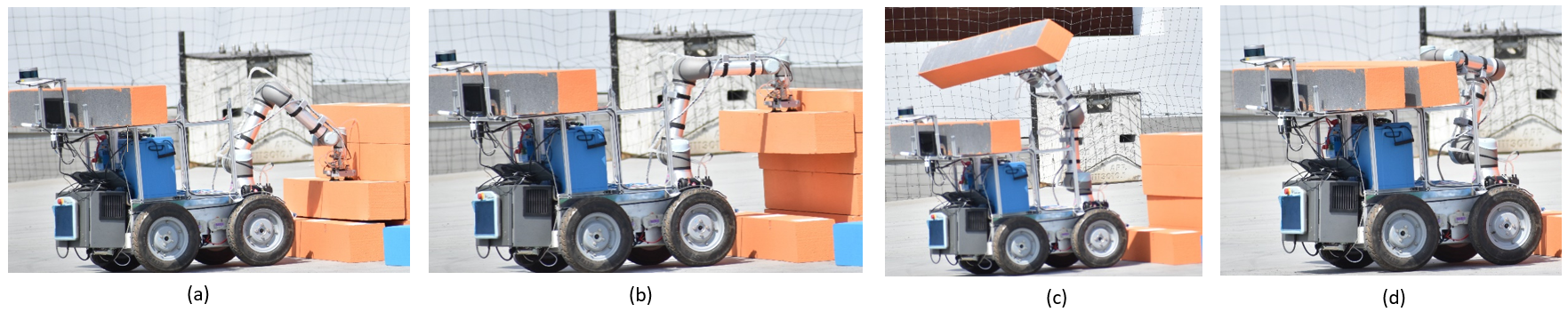}
\caption{Proposed algorithm in action: Accurate Pose estimation of cuboidal object and motion planning during Mohamed bin Zayed International Robotics Challenge (MBZIRC) 2020 \cite{MBZIRC-2020}} \label{fig:picking_result}
\end{figure*}
\subsection{Pose Estimation}\label{section:pose_est}
\textit{Super4PCS} \cite{mellado2014super} is a global point cloud registration algorithm which unlike \textit{4PCS} \cite{aiger20084} (four point congruent set) runs in optimal linear time and also memory efficient. Local registration algorithms like ICP guarantees convergence when the point cloud pairs are roughly aligned. Whereas in global registration methods such as 4PCS and Super4PCS, point cloud pairs can start in an arbitrary initial pose.
Super4PCS implements a smart indexing data structure to solve the core instance problem, i.e., estimating all point pairs found within a distance range of $(r-\epsilon, r+\epsilon)$. The algorithm handles angular and spatial queries with the ability of its data structure to extend to a higher dimension, allowing direct integration of auxiliary surface information (e.g. colour, normals) whenever available. In Fig \ref{fig:correction} (a), an artificial point cloud with its origin at \((0,0,0)\), which is the same as camera's origin, is transformed using resulting pose from Super4PCS given by equation \ref{eq:pose_mat} to align with target point cloud. The average time taken for point cloud registration using Super4PCS on planar surfaces of cuboidal objects during multiple trials in sunny and low light conditions is \(0.54s\) with an average rotational and translation error of \(\pm3.3^{\circ}\) and \(5.3 mm\) respectively.

\begin{equation} \label{eq:pose_mat}
Pose = 
\begin{bmatrix}
r_{11} & r_{12} & r_{13} & t_x\\
r_{21} & r_{22} & r_{23} & t_y\\
r_{31} & r_{32} & r_{33} & t_z\\
0 & 0 & 0 & 1
\end{bmatrix} 
\end{equation}

\subsection{Pose Error Estimation and Correction}
\label{section:error_estimation_correction}

The varying lighting condition and strong reflection could lead to the erroneous point cloud. The re-sampling and outlier removal algorithm applied during point cloud processing minimizes the possibility of an error during registration but could not guarantee precise results in case of a lack of information from the beginning itself. Any plane segment with minor loss of data points could still maintain its dimension information and can be filtered through the bounding box dimension filtering process. Such cases could lead to erroneous registration results. The error estimation and correction process in the algorithm was first applied to 3D rotation, followed by 3D translation correction over the corrected rotation. \par

\begin{figure}[ht]
    \centering
    \includegraphics[width=0.45 \textwidth]{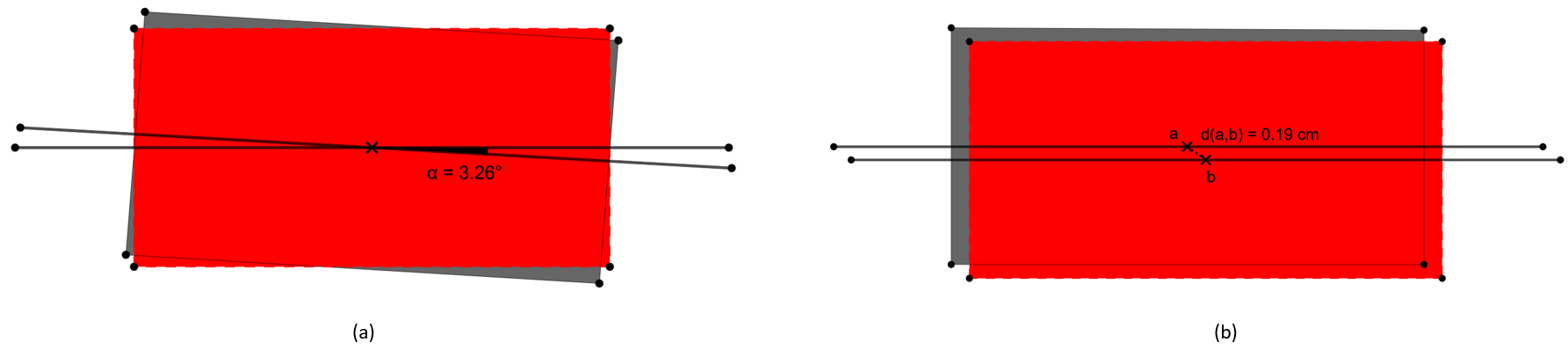}
    \caption{(a) Rotational error of $3.26^{\circ}$ between artificial point cloud (red) and target point cloud (gray). (b) Transitional error of 0.19cm between artificial point cloud (red) and target point cloud (gray) after compensating rotational error.}
    \label{fig:error_ilustration}
\end{figure}

\textit{Rotational error correction: } Compensating for the rotational error in the pose of a cuboidal object requires calculating the angle difference in the yaw between the target point cloud and the superimposed artificial point cloud Fig. \ref{fig:error_ilustration}(a). This is achieved by selecting two points on both target and artificial point cloud such that if the registration is perfect, then the angle difference between the line segments joining each pair of points results in zero. 

\begin{figure}
\centering
\includegraphics[width=0.45\textwidth, keepaspectratio]{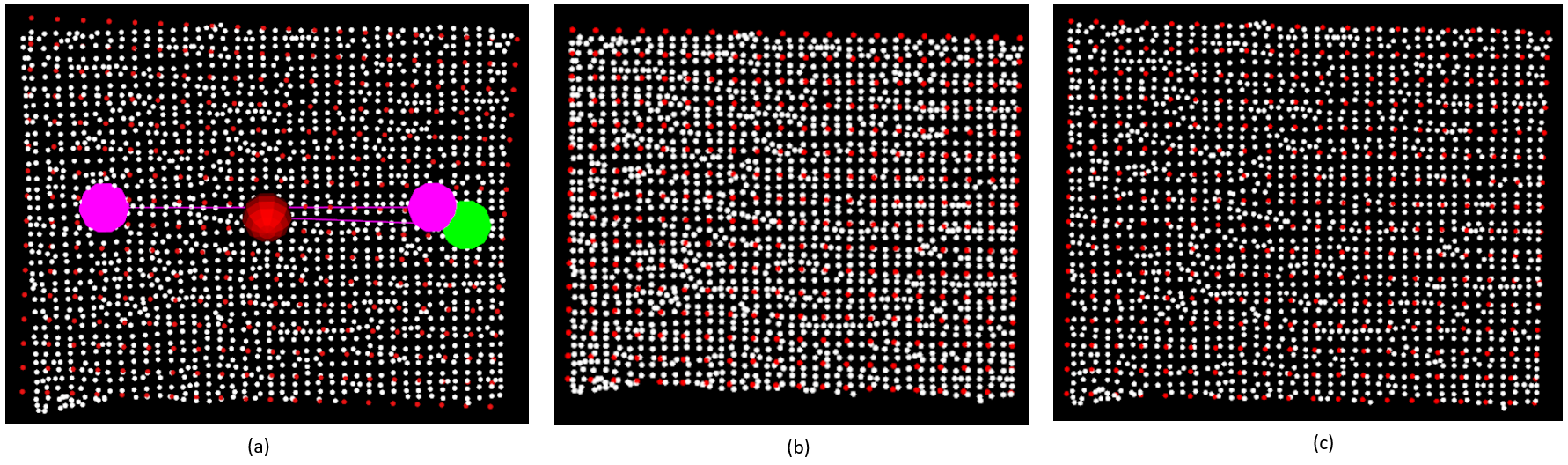}
\caption{(a) Artificial point cloud (red) is registered on target point cloud (white) with rotational error of 2.47\degree in yaw and translation error of 0.8 mm in x, 3.1 mm in y and -0.2 mm in z  (b) Rotational error compensated (c) Followed by translation error correction and giving final accurate pose with 0.23\degree accuracy in rotation and 0.3mm in transition (Euclidean distance from centroid)} \label{fig:correction}
\end{figure}

For the target point cloud, two points equidistant from the centroid and parallel to the larger sides of ROI are selected. Since the target point cloud is in the camera frame, the coordinate of these points is selected by either transforming the point along the major axis of the point cloud or the points \((T1, T2)\) shown in Fig \ref{fig:rgb_process}(c) are inversely projected using equations \ref{eq:inverse_proj_x}, \ref{eq:inverse_proj_y} and \ref{eq:inverse_proj_z}. From Fig. \ref{fig:correction} (a), \textit{T1} and \textit{T2} are represented as points in pink sphere.\par

For the artificial point cloud which is created with same dimensions as of target point cloud with its origin at \((0,0,0)\), the reference point for yaw correction is chosen such that first point \textit{A1} is the origin itself and other point \textit{A2} is along the major axis at \((x,0,0)\) (where \(x \in (0, artificial\_pointcloud\_width/2\)). \textit{A1}, \textit{A2} is then transformed using resulting pose from Super4PCS as shown in Fig. \ref{fig:correction} (a), as points in red and green sphere respectively. The line segment between \textit{T1}, \textit{T2} and between \textit{A1}, \textit{A2} as shown in Fig. \ref{fig:correction} (a), should ideally be parallel to each other if registration is perfect.In case of any rotational error, the angle between these two line segments is used to correct the resulting pose of Super4PCS by transforming the pose using \(R_{z}\) rotational matrix along the normal of the target point cloud segment.\par

\textit{Translation error correction: } Translation error is computed by comparing the target and artificial point cloud centroid Fig. \ref{fig:error_ilustration}(b). The centroid \textit{C1} of the target point cloud is the midpoint of the line segment joining points T1 and T2, as \(C1 = (t_{target}'(1), t_{target}'(2), t_{target}'(3))\). Whereas for artificial point cloud, its centroid \(C2 = (t_{artificial}'(1), t_{artificial}'(2), t_{artificial}'(3))\). The difference between \textit{C1} and \textit{C2} represents the translation error.\par

The estimated error in translation and rotation between the target and artificial point cloud is corrected by transforming the pose (equation \ref{eq:pose_mat}) with computed rotational error along z-axis ($\theta$ with corresponding rotation matrix $R_z(\theta)$) and translation error in $\delta t$ direction . As stated in equation \ref{eq:final_pose}, the final corrected pose is used for high-precision motion planning to grab the target object. The Fig \ref{fig:correction} shows the process of error correction in which Fig \ref{fig:correction} (a) shows the possibility of inaccurate registration resulting from Super4PCS. Using above stated method for rotational correction Fig \ref{fig:correction}(b) represents pose with rotational error eliminated and Fig \ref{fig:correction}(c) further removes any translation error and providing high precision pose.\par
\vspace{-5pt}
\begin{equation} \label{eq:final_pose}
\footnotesize
 FinalPose = Pose \times \begin{bmatrix}
R_{z}(11) & R_{z}(12) & R_{z}(13) & \delta t^T(1)\\
R_{z}(21) & R_{z}(22) & R_{z}(23) & \delta t^T(2)\\
R_{z}(31) & R_{z}(32) & R_{z}(33) & \delta t^T(3)\\
0 & 0 & 0 & 1
\end{bmatrix} 
\end{equation}
\vspace{-5pt}
\section{Results} \label{section: results}
In this section, we compare the pose accuracy and processing time of our approach for pose correction with ICP. When a globally registered point cloud is corrected for any pose error with the ICP algorithm, the resulting correction applied is dependent on the data points of the target point cloud, i.e., if the target point cloud has any inconsistency like missing corner data points, then it could introduce minor registration errors. In contrast, our approach uses geometrical information of the target area from its RGB and Depth image. We introduce our data point on the target point cloud to determine and correct pose accurately, as shown in Fig \ref{fig:correction}. Since our approach computes transitional and rotational error in constant time and for correction, it takes $\mathcal{O}(n)$ to transform the point cloud, unlike ICP, which takes $\Omega(n\log n)$) overall (where n is the number of data points), thus it can be concluded that our approach not only gets accurate pose but also saves execution time when compared to ICP as shown in Table 1. The achieved results shown in following table are from the experiment on the dataset captured during MBZIRC 2020 \cite{MBZIRC-2020} in which the average rotational error in pose estimation of bricks (cuboidal objects) before the correction was 3\degree \ in yaw and average transitional error of 3mm (Euclidean distance) before correction from the centroid of a target point cloud (Figure \ref{fig:picking_result}).

\begin{center}
\begin{table}
\renewcommand{\arraystretch}{1.2}
\label{table:result_table}
\centering
\begin{tabular}{ | M{7em} | M{6em}| M{6em} | M{6em} |} 
 \hline
 Algorithm & Avg. Execution Time (milliseconds) & Avg. Rotational Error after correction (degree) & Avg. Transitional Error after correction (mm) \\
  \hline
 ICP & 12 & 1.3\degree & 1.8 \\ 
 \hline
  Our approach & 0.21 & 0.2\degree &  0.4\\ 
 \hline
\end{tabular}
\caption{Performance comparison between ICP and our approach}
\end{table}
\end{center}
\vspace{-30pt}
\section{Conclusion}
This paper proposes an autonomous picking solution for cuboidal objects from a pile. Conventionally pose estimation is achieved by methods that align a reference point cloud with the target point cloud using model fitting or global point cloud registration algorithms. To compensate for added execution time and uncertainty in the pose accuracy of the ICP algorithm, we propose a simpler yet efficient way of computing the pose error and correcting it in linear time by generalizing a point selection process that depends on the geometry of the target object and its reference point cloud. This approach was tested and validated during the brick picking challenge in MBZIRC 2020.


\section*{Acknowledgment}

This work was a collaborative effort of Tata Consultancy Services(TCS) Research and Innovation Labs with the Indian Institute of Science, Bangalore(IISc), for the Mohamed bin Zayed International Robotics Challenge (MBZIRC).

\bibliographystyle{IEEEtran}
\bibliography{references}

\end{document}